\documentclass{article}

\usepackage[position, final]{neurips_2026}

\usepackage[utf8]{inputenc} % allow utf-8 input
\usepackage[T1]{fontenc}    % use 8-bit T1 fonts
\usepackage{hyperref}       % hyperlinks
\usepackage{url}            % simple URL typesetting
\usepackage{booktabs}       % professional-quality tables
\usepackage{amsfonts}       % blackboard math symbols
\usepackage{array}          % for \extrarowheight
\usepackage{nicefrac}       % compact symbols for 1/2, etc.
\usepackage{microtype}      % microtypography
\usepackage[table]{xcolor}
\usepackage{graphicx}
\graphicspath{{../}}
\hypersetup{hidelinks}
\definecolor{revisionblue}{RGB}{0,70,160}

\usepackage{amsmath}
\usepackage{amssymb}
\usepackage{mathtools}
\usepackage{amsthm}
\usepackage{arydshln}
\usepackage{enumitem}
\usepackage{wrapfig}
\newcommand{\mdp}{\mathcal{M}}
\newcommand{\expect}{\mathbb{E}}
\newcommand{\transition}{P_\mathcal{T}}

\title{{Toward Physical Recursive Self-Improvement: A Call for Body-Control Co-Design for Humanoid Robot}}

\author{%
Guiliang Liu$^{1,3}$, Bo Yue$^{1}$, Yi Jin Kim$^{2}$, \textbf{Kui Jia}$^{1,2}$
\thanks{Correspondence to: Kui Jia <kuijia@cuhk.edu.cn>.}\\
  $^{1}$The Chinese University of Hong Kong, Shenzhen,
  $^{2}$DexForce Co., Ltd,\\
  $^{3}$Shenzhen Loop Area Institute.\\
  \texttt{liuguiliang@cuhk.edu.cn, kuijia@cuhk.edu.cn} \\
}

\begin{document}

\maketitle

% \vspace{-0.2in}
\begin{abstract}
% \vspace{-0.05in}
Humanoid robots, as general-purpose physical agents, must integrate both intelligent control and adaptive morphology to operate effectively in diverse real-world environments. {While recent research primarily optimizes control policies for fixed bodies, we advocate body-control co-design as a path toward physical Recursive Self-Improvement (RSI): using experience from a humanoid's current body and policy to guide successive improvements to both.} 
Inspired by biological evolution, this approach enables robots to recursively adapt both their form and behavior to optimize performance within task-specific and resource-constrained contexts. 
Despite its promise, co-design in humanoid robotics remains a relatively underexplored domain, raising fundamental questions about its feasibility and necessity in achieving true embodied intelligence.
To address these challenges, {we discuss potential co-design methodologies grounded in strategic exploration, Sim2Real transfer, and meta-policy learning.} 
We further argue for the essential role of co-design by analyzing it from methodological, application-driven, and community-oriented perspectives.
Striving to guide and inspire future studies, we present open research questions, spanning from short-term innovations to long-term goals.
Ultimately, co-design offers a path toward physical RSI in humanoids that improves their bodies and policies while learning to make this improvement process more effective.
\end{abstract}

% \vspace{-0.15in}
\section{Introduction}
% \vspace{-0.05in}
As an emerging research area, Embodied AI posits that intelligence stems from an agent’s ability to actively explore, interact with, and learn from its environment in a continuous and dynamic manner. Within this learning paradigm, recent studies have developed various robot control models based on deep neural network backbones, enabling scalability across diverse tasks and environments~\cite{Zhao2023aloha,Lin2024VILA,liu2025rdtb,Ghosh2024Octo}.

In studies of embodied robot agents, their skills are closely tied to the physical form. For example, robot arms, grippers, and dexterous hands are commonly employed for manipulation tasks such as grasping, placing, and assembling objects~\cite{luo2023fmb,luo2024precise}. Similarly, wheeled robots, bipedal robots, and quadrupedal robots are designed for locomotion tasks, including walking, climbing, and navigation~\cite{xu2024human}. 
To develop general-purpose robots, recent studies have focused on humanoid robots. Equipped with dual arms, legged bodies, and advanced sensors, humanoid robots are well-suited for a wide range of mobile manipulation tasks under different environments~\cite{fu2024humanplus,Cheng2024Exbody,he2024omnih2o,He2024HOVER}. 

In recent years, the development of humanoid robots has primarily centered on control policy design, typically built upon predefined physical structures. 
These robotic designs are often the result of manual engineering and domain-specific heuristics, which are rarely subject to optimization within the embodied humanoid system.
However, embodied intelligence is not solely determined by control performance, but is also fundamentally grounded in agents' physical structure~\cite{gupta2021embodied}.
For instance, in natural systems, organisms evolve their body morphology to adapt to changing environmental conditions. Similarly, embodied agents should incorporate evolutionary mechanisms to adapt to task requirements and environmental dynamics.
{For humanoids, this coupling is particularly consequential because compatibility with human infrastructure constrains body dimensions, while manipulation and locomotion impose coupled balance and actuation demands. Changes to morphology therefore affect not only task performance but also hardware integration, reliability, and safety validation.}

% We view this coupled development as a path toward physical recursive self-improvement (physical RSI). Policy learning reveals the capabilities and limitations of the current body, informing design changes that reshape subsequent learning. A stronger recursive loop would also use accumulated experience to improve the design models, search strategies, or policy-adaptation procedures themselves. Thus, co-design provides a mechanism for pursuing physical RSI, rather than evidence that autonomous recursive improvement has already been achieved.

We view this coupled development as a path toward physical recursive self-improvement (physical RSI). Policy learning reveals the capabilities and limitations of the current body, guiding design changes that enable further learning. Accumulated experience could also improve the design models, search strategies, and policy-adaptation procedures that drive this cycle. For example, outcomes from earlier body–policy pairs can help predict which morphological changes will support better policy learning, guiding subsequent search. Co-design thus offers a starting point for physical RSI, with the longer-term goal of improving both the robot and its improvement process.

Specifically, to realize the physical RSI, robotic co-design jointly optimizes both the control policy and the morphological design of robotic systems~\cite{Carlone2019CoDesign,wang2025embodied}. 
While prior studies have explored co-design in quadruped robots~\cite{Belmonte2022RL4Leggy,Bjelonic2023CoDesignQuad}, soft robots~\cite{Wang2023PreCo}, bi-piedal robots~\cite{Cheng2024SERL,Ghansah2023HumanoidCoDesign} and modular robots~\cite{Zhao2020RoboGrammar,Whitman2020RLDesign} 
% (see Table~\ref{tab:summary-co-design})
, its extension to more advanced humanoid robots and connection to embodied intelligence remains largely unexplored.
It remains unclear how to efficiently discover the optimal design of a generalist humanoid robot capable of performing a variety of tasks. More importantly, the necessity of addressing such co-design problems in the development of embodied humanoid robots has yet to be fully established.

\begin{figure*}
    \centering
    \includegraphics[width=1.0\linewidth]{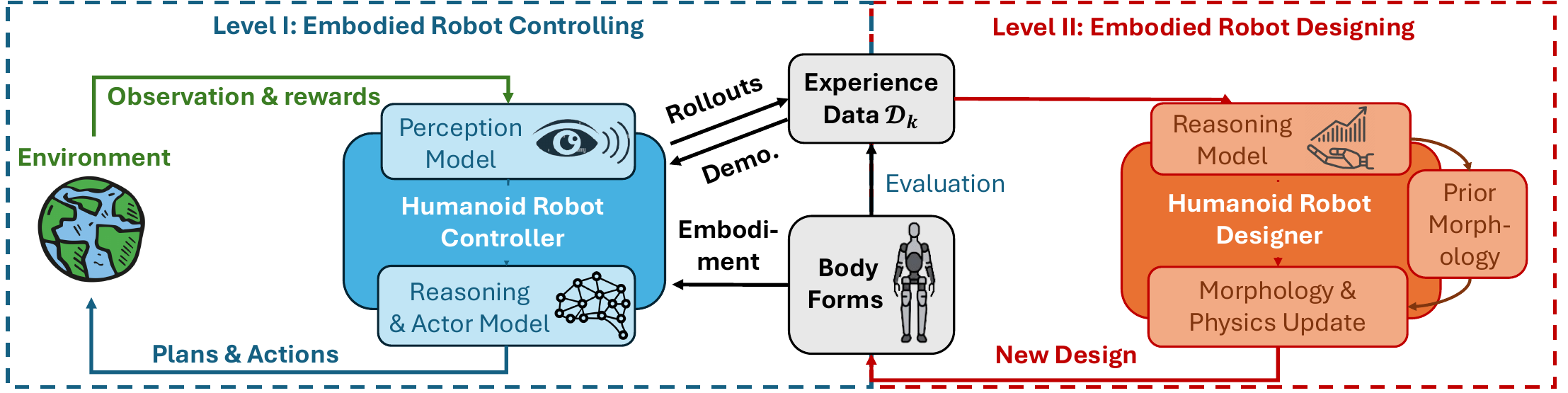}
    \caption{The co-design framework for humanoid robots, which can be formulated as a bi-level optimization problem, consisting of two interconnected phases: 1) learning the control policy for the humanoid robot, and 2) designing the robot's physical structure (Section~\ref{subsec:co-design-formulation}). As a potential mechanism for physical RSI, across design cycles, accumulated experience $\mathcal{D}_k$ can improve predictions of which body changes will benefit policy learning, making subsequent search more efficient.}\label{co-design_framework}
    \vspace{-0.2in}
\end{figure*}

\textbf{This article argues that humanoid body-control co-design provides a practical starting point for studying physical RSI, particularly when embodiment limits the capabilities achievable through policy optimization alone.}
As Figure~\ref{co-design_framework} shows, we formulate the humanoid co-design problem as a bi-level optimization. Such an optimizer can be integrated into the reasoning–acting architecture of an advanced controlling model, enabling an embodied humanoid robot to exhibit dexterity, mobility, perception, and intelligence.
{Our position does not imply that fixed-body humanoids are insufficient for all applications; rather, co-design becomes particularly important when controller optimization alone cannot overcome physical constraints. 
% We envision design-time hardware iterations, with a controller trained or adapted for each candidate body before physical validation.
}

Beyond the proposed formulation, we investigate an alternative perspective for realizing embodied humanoid robots based on predefined and manually specified designs.
We analyze why such paradigms prevail in recent humanoid robotics research and examine the potential challenges of adopting co-design, particularly regarding algorithmic complexity, physical evaluation, and design scalability.
To address these challenges, we introduce advancements in learning-based solvers, such as strategic robot structure exploration, the Sim2Real learning paradigm, and meta control policy, highlighting the feasibility of evolving humanoid robot architectures.

To understand the necessity of humanoid robot co-design, we investigate its unique advantages in facilitating robot morphology optimization, real-world task adaptation, and cross-disciplinary collaboration, examined from the perspectives of methodology, application, and community.
To realize these key advantages, we identify open questions within the humanoid robot co-design problem, highlighting those that may be tractable with current methodologies in the short term, as well as those that may depend on long-term advances in emerging research domains.
\vspace{-0.05in}
\section{Embodied Humanoid Robots}
\vspace{-0.1in}
\subsection{Architecture of Humanoid Robot}
\vspace{-0.1in}
Humanoid robots are a specialized type of physical robot designed to replicate human-like functionality~\cite{Saeedvand2019survey}. 
An ideal humanoid robot often has leggy designs in its {\it lower body}, featuring a {\it bi-pedal structure} that enables finishing locomotion tasks like walking, running, and maintaining balance.
The {\it upper body} includes {\it dual arms} equipped with dexterous hands as end-effectors, allowing the robot to perform complex tasks that require precise manipulation and human-like hand movements.
Their {\it sensor systems} commonly provide both {\it proprioception} and {\it exteroception}. Proprioception ensures internal body awareness by monitoring joint positions, angular velocity, and pose estimation, while exteroception enables perception of external states, such as LiDAR sweeps and RGB-D data.
% from depth cameras.

While such designs are ideal, building and fine-tuning a humanoid robot's architecture typically demands substantial effort and resources. This often necessitates certain simplifications, such as 
1) omitting end effectors in the dual arms,
2) constructing child-sized robots instead of full-sized ones,
3) utilizing a wheeled base for the lower body, and
4) downplaying the amounts and quality of sensors.
Figure~\ref{fig:humanoid-examples} provides illustrative examples of humanoid models.

\noindent\textbf{Desiderata for Humanoid Robot.} Given their human-like body structure, a fundamental requirement for humanoid robots is the ability to operate seamlessly within human environments. This enables them to collaborate closely with humans or take on dangerous or physically demanding tasks. During operation, the robot should exhibit natural behavior, adhering closely to human behavioral norms, even when performing long-term tasks across varying environments.
These desiderata demand a control model with strong generalizability and adaptability, which traditional optimization-based control methods often struggle to achieve. Essentially, fulfilling this requirement calls for the development of embodied intelligence in humanoid robots.
\vspace{-0.15in}
\subsection{Embodied Intelligence in Humanoid Robot}\label{subsec:embodied-humanoid-robot}
\vspace{-0.1in}
Unlike traditional approaches that passively learn from fixed datasets, Embodied AI requires agents to actively explore, interact with, and learn from their environment in a continuous and dynamic manner.
Specifically, to enable the learning of an embodied humanoid robot, it often requires the robot to have four key abilities, including:
1) {\bf Dexterity}: the ability to manipulate various objects with precision, delicacy, and intricacy.
2) {\bf Mobility}: the capability to move and navigate through environments with different terrains and conditions.
3) {\bf Perception}: the skill to gather, interpret, and understand environmental information from sensors.
4) {\bf Intelligence}: the ability to process information, reason about sub-goals related to a given task, and adapt to diverse tasks and environments.

These capabilities reflect not just the functionality of a humanoid robot in specific tasks like locomotion or manipulation but also emphasize generalization to a wide range of real-world scenarios, thereby advancing toward zero-shot deployment of humanoid robots for realistic applications.

\vspace{-0.15in}
\subsection{Implementation for Embodied Humanoid Robot.}
\vspace{-0.1in}

To achieve the above abilities, recent studies have implemented embodied humanoid robots using a two-layer architecture consisting of a high-level reasoning model and a low-level action model~\cite{team2025gemini,yuan2025being,figure2025helix}. This hierarchical design is inspired by the functional organization of the human brain, where the cerebrum is responsible for logical reasoning and decision-making, while the cerebellum governs fine-grained motor control and coordination, detailed as follows:

\textbf{Humanoid Robot Reasoning Model.} The reasoning model is typically implemented as a large-scale robotics foundation model designed to perform logical reasoning over the necessary steps for a robot to complete given tasks. The model takes as input natural language instructions describing the task, along with perceptual data, primarily visual signals collected from the environment. Its outputs consist of high-level plans to guide robotic execution. These may include sub-task descriptions and intermediate goals, as well as more detailed robotic outputs such as motion trajectories, grasp poses, and contact points (e.g., affordances). To learn this reasoning model, the training of the robotic reasoning model is typically achieved by fine-tuning a pre-existing Vision-Language Model (VLM) using refined robotic operation data, which is either collected through teleoperation or generated using synthetic data engines.

\textbf{Humanoid Robot Action Model.} 
The action model takes the outputs from the reasoning model (either as explicit data or implicit latent variables) as goals $g\in\mathcal{G}$ and predicts the corresponding control signals, such as joint angles or torques for the robot's joints at each time step. To learn action policies, previous methods commonly formulate the learning environment as a (partially observable) Markov Decision Process (MDP). $\mdp=(\mathcal{S}, \mathcal{A}, \mathcal{O}, P_\mathcal{T}, r,  \mu_0, \gamma)$, where:
1) Within the state space $\mathcal{S}$, a state $s\in\mathcal{S}$ records the complete environmental information and the robot's internal states. 
2) $\mathcal{A}$ denotes the action space, and action $a \in \mathcal{A}$ denotes the angles or torques at joints of the humanoid robot.
3) $o\in\mathcal{O}$ denotes the observations obtained from the robot's sensors, encompassing both proprioceptive inputs that reflect the humanoid’s internal state and exteroceptive inputs that capture information about the external environment.
% By incorporating the temporal observations
% \(\boldsymbol{o}_{t}^{H} = [\boldsymbol{o}_{t},\boldsymbol{o}_{t - 1},\cdots,\boldsymbol{o}_{t - H}]\), 
% the control model can summarize a state as $s_t=[\boldsymbol{o}_{t}^H,\mathbb{P}_t]$, where $\mathbb{P}_t$ is the privileged information which the robot can't access in realistic deployment, including base velocity $v_t$, terrain height, external disturbance and ZMP features. 
4) $r$ denotes the reward functions, which typically consist of penalty, regularization, and task rewards. In particular, the magnitude of the task reward should closely reflect how well the robot accomplishes the given goal $g$.
% These reward signals significantly influence the optimality of the control policy, for which we provide a detailed introduction in the following.
5) $\transition\in \Delta^{\mathcal{S}}_{\mathcal{S}\times\mathcal{A}}$ denotes the transition function as a mapping from state-action pairs to a distribution of future states. 
6) $\mu_0\in\Delta^{\mathcal{S}}$ denotes the initial state distribution.
7) $\gamma\in(0,1]$ denotes the discount factor. 
% For brevity, we denote $\mathcal{M}_{/R}$ to denote the MDP without knowing the reward. 
% In this work, we mainly study the episodic MDPs where 
% The planning stops at a terminating state $\Tilde{s}$, and the corresponding terminating time is denoted as $T\in(0,\infty)$.

Under this MDP, the humanoid action model can be represented as a meta policy $\pi\in\Delta^{\mathcal{A}}_\mathcal{S\times G}$ that can scale to diverse goals $g\in\mathcal{G}$ under different environmental states $s\in \mathcal{S}$. During training, the goal is to maximize the expected cumulative discounted rewards:
\begin{align}
    \max_{\pi\in \Pi}\mathcal{J}(\pi,\mdp)=\max_{\pi\in \Pi}\expect_{\mu_0,p_\mathcal{T},\pi}\left[\sum_{t=0}^{\infty}\gamma^tr(s_t,a_t,g)\right]\label{obj:rl}
\end{align}
\vspace{-0.05in}
\section{Position Proposal and Alternative Views}
\vspace{-0.05in}
In the following sections, we introduce a co-design framework that jointly updates control policies and humanoid morphology. Additionally, we present an alternative perspective to our proposal.
% : embodied intelligence should be grounded in predefined humanoid structures without evolution.
\vspace{-0.1in}
\subsection{{A New Perspective: Body-Control Co-Design as a Mechanism for Physical RSI}}\label{subsec:co-design-formulation}
\vspace{-0.1in}
In natural environments, animals exhibit remarkable embodied intelligence, leveraging their evolved morphologies to learn and perform complex tasks~\cite{gupta2021embodied}. 
Inspired by this, {\it we argue that evolutionary principles should guide the recursive improvement of both morphology (physical form) and control in embodied humanoid robots.}
While prior research commonly focused on perception, reasoning, and control within fixed robot structures, our position is that the robot's physical form itself should also be subject to optimization as a core component of its design. The simultaneous optimization of a humanoid robot’s action model~\footnote{Since compared to robot acting, task reasoning is typically less dependent on the robot’s detailed physical structure and updating a VLM is often costly, the reasoning model is generally not updated alongside changes to the robot’s morphology.} and physical components can be formulated as a co-design problem, integrating both control and morphology in the design process.

As illustrated in Figure~{\ref{co-design_framework}}, the robot co-design problem requires the joint optimization of both control policies and physical robot modules to maximize overall performance, while adhering to resource constraints such as cost~\cite{Carlone2019CoDesign}. 
Experience accumulated across iterations can also refine design models, search strategies, and policy adaptation, providing a path toward physical RSI.
For example, evaluations of previous body–policy pairs can update a learned design model that proposes subsequent morphologies, allowing the design process itself to improve across cycles.
When extending this framework to a learning-based setting, the co-design task typically involves {an inner loop for training the control policy and an outer loop for updating the robot's physical parameters; the latter need not use gradients or RL.}

% Based on an POMDP $\mathcal{M}(\psi)$, 
{Specifically, during controller optimization,} we use the RL algorithm to optimize the goal-aware policy function by maximizing $\mathcal{J}(\pi,\mdp_\psi)$ in the objective~(\ref{obj:rl}), where the configuration of this learning environment (i.e., MDP) depends on a specific {physical} robot structure, denoted by $\psi \in \Psi$.
{In principle, $\psi$ can include high-level choices such as topology and actuator or sensor allocation, mid-level parameters such as link geometry, inertia, transmissions, and compliance, and detailed engineering choices such as cable routing, electronics, and thermal management. Near-term studies can restrict optimization to simulator-compatible parameters while keeping topology fixed. Extending beyond this scope requires integration with CAD, manufacturing constraints, and mechanical and electronic models that are not fully represented in standard robot-learning simulators.}
% where the expectation is taken over the distribution of goals $p_{\mathcal{G}}$, the policy $\pi$, and the probabilistic transition dynamics and initial state distribution $(\transition, \mu_0)$, conditioning on the MDP associated with the given physical physical robot structure $\mathcal{M}(\psi)$.

Based on the policy performance, {we update the robot's structure in the outer loop}, which necessitates formulating humanoid robot co-design as a bi-level optimization problem. Moreover, the design often incorporates system-level constraints, which define strict requirements for the desired system behavior (e.g., resembling human behavior) or impose limitations on the resources (e.g., costs of robot modules)~\cite{Carlone2019CoDesign}. The optimization problem can be described as:
\begin{align}
    \max_{\psi\in\Psi}\max_{\pi\in\Pi}
    \mathcal{J}(\pi,\mathcal{M}_\psi\mid\mathcal{D}_k)
    \quad \text{s.t. } f_c(\psi)\leq\epsilon.
    \label{obj:bi-level}
\end{align}
Here, \(\mathcal{D}_k\) denotes experience accumulated in previous body–policy evaluations, thereby guiding subsequent optimization during physical RSI.
% Physical RSI further requires this experience to improve the design models, search strategies, or policy-adaptation procedures themselves.
For multi-task co-design, we interpret $\mathcal{J}$ as the expected return over a declared goal distribution $p_{\mathcal G}$. 
In practice, we can use controller rewards to guide the inner learning process, whereas the outer search can compare task success, robustness, energy, and component-cost estimates through explicit constraints or Pareto trade-offs. For example, success can be maximized subject to energy and hardware-cost budgets. Fabrication time can be included as an uncertain engineering estimate. 

% In practice, however, finite-budget policy training introduces noise and adaptation bias, requiring repeated evaluations and consistent training budgets.
% \rev{
% To this end, LLM-assisted reward construction and generative simulation can support the inner controller-learning process~\cite{ma2024eureka,ma2024dreureka,Wang2024RoboGen}.
% , but these enabling techniques should not be interpreted as complete body-control co-design frameworks.}

Under this perspective, each co-design iteration generates experience that informs the next body–policy pair. Equation~(\ref{obj:bi-level}) expresses this coupled optimization conditioned on accumulated experience \(\mathcal{D}_k\). Controller performance reveals embodiment-dependent limitations, while morphological updates reshape subsequent policy learning. Beyond improving bodies and policies, \(\mathcal{D}_k\) can also help refine the design models, search strategies, and adaptation procedures that guide this process, providing a path toward physical RSI. A key research question is therefore how experience across designs can make future search and adaptation more effective (Section~\ref{sec:open-questions}).

\vspace{-0.05in}
\subsection{Alternative Views: Intelligence Arises from Fixed Humanoid Robots Structure}\label{subsec:alternative}
\vspace{-0.1in}
While the co-design approach provides a significantly broader design space for enhancing the performance of humanoid robots, most recent studies, spanning manipulation~\cite{Ze2024Humanoid3DDiffusion,li2024okami,Cheng2024OpenTeleVision}, locomotion~\cite{Chen2024LipschitzConstrainedPolicies,Zhuang2024Parkour,Long2024PIM,Gu2024Locomotion}, and human motion imitation (i.e., teleoperation)~\cite{he2024h2o,fu2024humanplus,he2024omnih2o,He2024HOVER,Cheng2024Exbody,ji2024exbody2}, continue to focus on controlling fixed, pre-defined humanoid platforms, without considering structural modifications to the robot itself. 
Even with the recent surge in exploring embodied intelligence in humanoid robots across a wide range of everyday tasks~\cite{Nasiriany2024RoboCasa}, {\it robotic co-design, as an effective technique in robotics research, remains largely underexplored} in this context~\cite{tong2024advancements}. This trend reflects a prevailing assumption that {\it "The predefined and fixed physical structures are sufficient for supporting the development of humanoid robots"}.

This perspective essentially treats the robot’s structure as a fixed component of the environment's dynamics, which can be estimated and adapted to, but not actively optimized.
Conceptually, this assumption is prevalent in the RL literature~\cite{sutton2018reinforcement}, which serves as a foundational algorithm for learning-based control in humanoid robots and inherently shapes subsequent research directions.
{This fixed-body perspective has engineering advantages: standardized platforms support shared datasets, controller reuse, manufacturing scale, maintenance, and accumulated safety evidence. 
Modular end-effectors and task-specific tools may meet new requirements more economically than redesign. Co-design should therefore complement these approaches when their capability or cost limits become restrictive, while multi-task evaluation is needed to prevent morphological over-specialization.} 
More importantly, in practice, there are significant challenges associated with co-designing humanoid robots, further reinforcing the reliance on manually designed, fixed-structure humanoid platforms.
% To understand this phenomenon, in this section, we discuss the practical challenges associated with co-designing humanoid robots, thereby providing a justification for why previous works have predominantly relied on manually designed, fixed-structure humanoid platforms.
% \guiliang{maybe not just challenge, we must discuss why it is sufficient.}

\textbf{1) Complexity of the Co-design Problem.}
In addition to learning control policy, the co-design problem incorporates a second-level optimization loop for refining the robot’s physical design~\cite{Carlone2019CoDesign,gupta2021embodied,Sartore2023HumanoidCoDesign,Bjelonic2023CoDesignQuad,Cheng2024SERL,fadini2024making}. For humanoid robots, this process becomes especially challenging due to their complex upper and lower body structures, which often involve varying configurations of motors, joints, sensors, and body components. Consequently, exploring the high-dimensional design space and identifying optimal configurations demands substantial computational resources. In many cases, due to the intricate interdependencies between the design and control parameters, the bi-level optimization in the co-design problem may struggle to converge. Without additional restrictions, the optimal solution may not be uniquely identifiable or even computationally tractable.

\textbf{2) Difficulties in Physical Evaluation.} To evaluate the optimality of a humanoid robot structure, it is crucial to deploy the robot in task-specific scenarios and assess how effectively it can adapt its control model to complete those tasks. This process often requires modifying certain components of the robot based on the proposed design. However, unlike simpler robotic systems, humanoid robots have highly complex and interdependent structures~\cite{Katic2003Humanoid,Saeedvand2019survey,Darvish2023Humanoid}. Modifying one part frequently leads to changes in the robot’s overall physical configuration.
For instance, adjusting the length of the thighs affects the robot’s weight distribution and center of mass. These changes, in turn, influence both kinematic properties (e.g., motions and velocities) and dynamic characteristics (such as inertia and gravity models). The technical challenges in physical reconfiguration limit the feasibility of iterative structural design and evaluation in real-world applications.

\textbf{3) Limited Scalability Across Tasks.}
Robotic co-design typically aims to enhance performance for specific tasks~\cite{Carlone2019CoDesign}. 
For instance, \cite{Cheng2024SERL} optimized the leg length of a bipedal robot to achieve maximum walking velocity. 
Similarly, \cite{Bjelonic2023CoDesignQuad,chen2024pretraining} explored the joint optimization of mechanical structures and control policies to improve the locomotion capabilities of quadruped robots.
However, embodied humanoid robots, with physical structures resembling those of humans, are designed to generalize across a wide range of tasks and environments within human workspaces. This requires multi-dimensional capabilities, including dexterity, mobility, perception, and intelligence (Section~\ref{subsec:embodied-humanoid-robot}). The task-specific optimization frameworks commonly used in traditional robotic co-design cannot be directly applied to the inherently cross-task nature of humanoid robots. This lack of task scalability limits the overall utility of co-designed systems, particularly when targeting general-purpose humanoid platforms intended for reuse across diverse applications.

\vspace{-0.05in}
\section{Feasibility of Humanoid Robot Co-Design}
\vspace{-0.05in}
To address the inherent challenges of humanoid robot co-design, this section investigates its feasibility by proposing a set of potential solutions. In particular, we introduce three key strategies: strategic exploration, the Sim2Real paradigm, and meta-policy learning. These approaches are aimed at tackling critical issues in co-design, including the complexity of joint design and control, the challenges of physical evaluation, and the limited scalability across diverse tasks.
Moreover, by leveraging recent advances in control algorithms, simulated environments, foundation models, and decision-making policies, these strategies establish a robust foundation for the development of next-generation co-design algorithms for embodied humanoid systems.
{
To this end, Initial humanoid studies address actuator co-design for chin-up performance~\cite{jin2025evolutionary}, joint morphology-control optimization for fall recovery~\cite{yue2025toward}, and surrogate-guided locomotion co-adaptation~\cite{du2025efficient}. 
% These bounded settings motivate the agenda, but do not establish scalable multi-task co-design or reliable transfer across arbitrary bodies.
}
% \guiliang{we solve .. problem via .... add details, they form the important paradigm for humanoid co-design, maybe also add a plot.}
\vspace{-0.15in}
\subsection{Strategic Exploration under Constraints}\label{subsec:exploration}
\vspace{-0.1in}
In the co-design literature, genetic algorithms take a critical role in modifying robot structures via crossover, mutation, and replacement operations~\cite{gupta2021embodied,Belmonte2022RL4Leggy,Bjelonic2023CoDesignQuad,Cheng2024SERL,fadini2024making}.
{These methods combine stochastic variation with performance-based selection. Nevertheless, evaluating many candidate designs can remain expensive when informative priors or learned guidance are limited.}
This challenge becomes especially pronounced in unstructured and high-dimensional design spaces, which induces computational burden (see alternative views in Section~\ref{subsec:alternative}). 
% The computational burden of exploring vast, unconstrained morphologies without informative priors or guidance can significantly hinder the efficiency and scalability of the co-design process.

To address this limitation, {\it strategic exploration} has emerged as a promising approach for accelerating the search process. 
In the RL literature, a variety of algorithms have been developed to promote more efficient exploration~\cite{ladosz2022exploration}. For instance, recent work has proposed provably efficient strategies based on Bayesian updates~\cite{Balakrishnan2020BayesianExplore,mansour2022bayesian}, the Upper Confidence Bound (UCB)~\cite{Lu2019UCB}, and uncertainty-driven heuristics~\cite{Cai2020Provably,Jin2020Provably}, achieving significantly lower regret bounds compared to purely random exploration.
Motivated by these advances, we propose formulating the physical robot structure process as an MDP, which enables the application of strategic exploration techniques to more effectively explore the design space of humanoid robots.
{Model-based approaches offer a complementary way to reduce expensive evaluations: a learned surrogate can predict performance and uncertainty from design parameters, guiding the search toward promising candidates~\cite{du2025efficient}. Morphology-conditioned dynamics models instead predict how candidate bodies respond to actions and may support controller adaptation or local design sensitivity analysis. 
% Both require validation on unseen designs because inaccurate predictions can misdirect the search.
}

In addition to accelerating exploration, {another important strategy for reducing search complexity is to} {\it constrain the design space} with the following methods: 1) Rather than modifying the entire robot structure, {\it the exploration can be limited to a few key modules or components}. For instance, \cite{Ghansah2023HumanoidCoDesign, chen2024pretraining} focused on optimizing the thigh and shank lengths of humanoid robots. Similarly, \cite{Sartore2023HumanoidCoDesign} investigated the configuration of various motors and the inertial parameters of robot links. Other studies, such as \cite{bravo2024engineering, Bjelonic2023CoDesignQuad}, examined the parameters of parallel elastic knee joints.
{2) From a theoretical perspective, if the feasible design space is nonempty and compact and the optimized value function $V(\psi)=\max_{\pi\in\Pi}\mathcal{J}(\pi,\mathcal{M}_\psi)$ is well-defined and continuous, the Extreme Value Theorem~\cite{boyd2004convex} guarantees that $V$ attains a maximum\footnote{This is an existence statement, not a guarantee of uniqueness, computational tractability, or convergence of practical bi-level algorithms.}.
% Discrete hardware choices, simulator error, and noisy policy training require separate treatment.
}

By strategically exploring a compact and bounded design space focused on key modules of the humanoid robot, the computational burden of humanoid robot co-design can be significantly reduced. This approach substantially alleviates the complexity of the bi-level optimization (Section~\ref{subsec:alternative}).

\vspace{-0.15in}
\subsection{A Sim2Real Paradigm for Evaluation}\label{subsec:sim2real}
\vspace{-0.1in}
Guiding structural updates of humanoid robots typically requires evaluation signals from the current design. Traditional robot co-design studies often involve building physical hardware and assessing its performance in real-world tasks~\cite{Whitman2020RLDesign,Ghansah2023HumanoidCoDesign,Bjelonic2023CoDesignQuad,bravo2024engineering}. However, as discussed in alternative perspectives (Section~\ref{subsec:alternative}), {repeated real-world design and evaluation can be prohibitively expensive for humanoid robots due to the complexity of their physical structures.} 

{To reduce the number of real-world evaluations, an effective approach is to conduct initial design and evaluation within simulated environments.} For example, studies such as~\cite{Schaff2019RLDesign, Luck2019CoAdapt} and~\cite{Bjelonic2023CoDesignQuad, chen2024pretraining, Cheng2024SERL} use simulators such as MuJoCo and Isaac Gym to learn control policies and evaluate task performance across various physical robot structures. While simulation reduces the cost and complexity of building physical hardware, the Simulation-to-Reality (Sim2Real) gap remains a major challenge: a robot that performs well in simulation may not exhibit the same level of performance in the real world. Additionally, for humanoid robots with soft components, simulating their morphology, especially for contact-rich interactions, remains difficult~\cite{Schaff2022SoftRLDesign}.

As a result, developing a feasible Sim2Real paradigm for humanoid co-design requires effective strategies to bridge the gap between simulation and reality. Key approaches include:
1) Extending {\it domain randomization and domain adaptation} techniques to humanoid control environments, which remains a critical direction for future research~\cite{wilson2020survey}.
2) {\it Developing simulation platforms with high photorealism and physical fidelity} to minimize the impact of the Sim2Real gap. For instance, recently developed simulation environments and engines such as RoboCasa~\cite{Nasiriany2024RoboCasa}, Isaac Lab~\cite{mittal2023orbit}, MuJoCo-Playground~\cite{mujoco_playground_2025}, Genesis~\cite{Genesis}, and ManiSkill~\cite{Tao2024ManiSkill3} have demonstrated promising capabilities in accurately modeling real-world scenes and physical interactions.
3) Beyond explicitly modeling objects, scenarios, and physical laws, recent work has introduced the concept of the {\it World Function Model}~\cite{cong2025can, agarwal2025cosmos}. This paradigm treats simulation as a regression task, where the model predicts future states of the environment in response to perturbations (i.e., actions). This offers a more flexible and data-driven alternative to traditional simulators.
{However, higher visual or physical fidelity can not alone guarantee correct morphology-dependent behavior. Candidate designs should be screened using calibrated actuator and contact models as well as uncertainty in mass, friction, compliance, and latency, with sensitivity checks for simulator artifacts. Shortlisted designs still require component-level or hardware-in-the-loop testing before deployment.}
\vspace{-0.15in}
\subsection{Controlling Humanoid Robot via Meta Policy}\label{subsec:meta-policy}
\vspace{-0.1in}
As discussed in Section~\ref{subsec:alternative}, traditional approaches to robotic design and control have primarily focused on optimizing simple robots to complete specific tasks~\cite{Carlone2019CoDesign}. In contrast, embodied humanoid robots, as complex, legged, dual-arm systems equipped with numerous sensors, are designed to learn generalizable policies that can be applied across a wide range of tasks and environments. This fundamental difference renders previously proposed robot co-design methods not directly transferable to the problem of controlling embodied humanoid robots.

To enable the designed robot to perform effectively across multiple tasks, a critical approach is to learn a meta-policy $\pi$ that allows the humanoid robot to solve a wide range of everyday human tasks involving both manipulation and locomotion~\cite{Nasiriany2024RoboCasa}.
In the context of humanoid control, the meta-policy can be modeled as a goal-conditioned policy $\pi: \mathcal{S} \times \mathcal{G} \rightarrow \mathcal{A}$, where a goal $g \in \mathcal{G}$ may represent various forms of task specification, such as commands~\cite{Gu2024Locomotion, He2024HOVER, Zhuang2024Parkour}, target poses~\cite{he2024h2o, fu2024humanplus}, affordances~\cite{Lu2024MobileTeleVision, Zhang2024wococo}, or more generally, natural language descriptions of tasks~\cite{team2025gemini, figure2025helix, yuan2025being}.
Recent advances in generalizable decision models, such as decision transformers~\cite{Chen2021DT} {and action-chunking transformers~\cite{Zhao2023aloha}}, provide effective backbone architectures for implementing such meta-policies, enabling flexible and scalable control across a wide range of goal specifications.

To integrate robotic co-design into policy learning, goals can be randomly sampled from a predefined goal pool during training. The robot's policy is then conditioned on each goal and adapted according to a proposed design configuration $\psi$. Similarly, during evaluation, the effectiveness of a given physical robot structure can be assessed by measuring how well it facilitates the learning of a policy that successfully achieves a diverse set of goals across varying environments.
{A goal-conditioned policy is not automatically morphology-aware: body changes alter dynamics and sensing and may also change observation and action dimensions. Reliable transfer to unseen designs, limited-budget adaptation, and avoiding negative transfer therefore remain open challenges.}
{The potential benefit is amortized controller learning: a policy trained across morphologies can initialize or adapt controllers for new candidates, reducing the need to train every design from scratch. 
% However, comparisons must account for this pretraining cost and distinguish genuinely poor designs from designs whose controllers have not yet adapted.
}

\vspace{-0.15in}
\section{Necessity of Co-Designing Humanoid Robot}
\vspace{-0.1in}
In the previous section, we explored the feasibility of jointly modeling the robot's physical structure and its control policy, outlining key strategies that make such co-design tractable. In this section, we go a step further and argue for the necessity of co-design in the development of embodied humanoid robots from the perspective of methodology, application, and community.
\vspace{-0.1in}
\subsection{Principled  Optimization of Robot Morphology}
\vspace{-0.1in}
While significant progress has been achieved using pre-designed humanoid robots in both locomotion and manipulation tasks, there remains a lack of principled methods for evaluating the optimality of these designs. In practice, to determine robot morphology in Figure~\ref{fig:humanoid-examples}, {\it it commonly relies on engineers’ intuition and experience, rather than through systematic optimization}.
Such practice is often inefficient, since the robot's physical design is independent of the training of its planning models and control policies. 
The full capabilities and limitations of a given design often only become apparent when other research groups attempt to tackle more complex locomotion or manipulation tasks, revealing shortcomings that were not initially evident. 
These insights are used retrospectively to inform the development of robots, which typically progresses slowly due to the lack of systematic design methodologies.
For example, in the initial design of the Unitree H1 robot, the limited degrees of freedom (DoFs) in its arms significantly constrained its ability to perform manipulation tasks that require rich interactions with objects. Recognizing this limitation, the developers addressed it in the subsequent version, Unitree H1-2~\cite{unitree2025H1}, by increasing the number of DoFs in each arm from 4 to 7. However, this design revision took over a year to implement.

In addition to improving the efficiency of robot development, {co-design} can significantly enhance its {\it efficacy}. From an algorithmic standpoint, allowing the robot's design $\psi$ to vary enables the learning algorithm to explore a larger joint search space over both morphology and control. This expanded space allows for the discovery of design-policy pairs that maximize overall performance:
\begin{equation}
    \max_{\psi \in \Psi} \max_{\pi \in \Pi} \mathcal{J}(\pi, \mathcal{M}_\psi) \geq \max_{\pi \in \Pi} \mathcal{J}(\pi, \mathcal{M}_{\psi'}), \quad \forall \psi' \in \Psi
\end{equation}
Here, $\mathcal{J}$ denotes the expected return of policy $\pi$ in the environment defined by the morphology $\mathcal{M}_\psi$. This inequality highlights the potential performance gains from jointly optimizing both the robot’s design and its control policy.
% \rev{The inequality follows from inclusion of the fixed-body baseline in the feasible joint search space and allows equality; it does not establish strict improvement or practical search efficiency. Evidence that body optimization is needed for a target capability should distinguish physical limits from insufficient controller training and compare strong fixed-body and tooling alternatives under comparable resources.}
Without {principled optimization}, manually identifying the optimal design $\psi^*\in\Psi$ is highly challenging, particularly for {humanoid robots} expected to perform {diverse tasks} and adapt to {complex, real-world environments} encountered in everyday scenarios.
\textit{{When physical design limits the required capabilities, we argue that humanoid robot co-design is an important route to systematically improving body-policy combinations}}.
\vspace{-0.1in}
\subsection{Adaptive Body Shaping for Real-World Tasks}
\vspace{-0.1in}
In practice, the specific requirements for a humanoid robot’s capabilities vary depending on the deployment environment of each real-world application.
For example, in industrial settings, humanoid robots are often tasked with manipulating a variety of objects for relocation, rearrangement, or assembly. In such scenarios, dexterity becomes a critical factor, as robots must precisely and efficiently handle components of different shapes, sizes, and material properties~\cite{Kuindersma2016Atlas,Long2024PIM,Gu2024Locomotion,Zhang2024WholebodyLocomotion,Chen2024LipschitzConstrainedPolicies}.
In contrast, when deployed as patrol robots in environments such as university campuses or public facilities, humanoid robots must robustly navigate to different locations under diverse terrains (e.g., slopes, stairs) and environmental disturbances (e.g., dynamic obstacles or weather conditions). In these applications, robustness and mobility become the key performance criteria~\cite{Smith2012DualArm,billard2019trends,li2024okami,Ze2024Humanoid3DDiffusion}.
Most importantly, {\it {there can be a design-dependent trade-off between dexterity and mobility}} in the design of a humanoid robot's body structure. Achieving dexterous manipulation typically requires highly flexible arms and delicate actuators. {Depending on actuator placement and link geometry, these choices can increase arm inertia or shift mass distribution, affecting balance and locomotion.
% the direction and magnitude must be evaluated for each design.
}

A similar trade-off observed in humanoid robots can also be found in human physiology. For example, the body composition of boxers and runners differs significantly in terms of muscle distribution and weight allocation. 
These athletes often dedicate substantial time to optimizing their bodies to enhance the specific skills required in their respective sports.
Just as athletes undergo intensive training camps to simultaneously develop both their physical form and technical skills, the co-design process of a humanoid robot’s body and control policy can be viewed as a training camp for humanoid robots. {Through design-time iterations of morphology and behavior, co-design can explore body-policy combinations suited to a declared family of tasks. 
% The athlete analogy illustrates compromises among capabilities, rather than suggesting that one body can be simultaneously optimal for all activities.
}
{\it {We argue that co-design is particularly valuable when the target applications expose physical limitations that controller alone cannot overcome.}}

\vspace{-0.15in}
\subsection{Fostering Cross-Disciplinary Collaboration}
\vspace{-0.1in}
Co-designing the control model and body structure of an embodied humanoid robot is fundamentally a multidisciplinary research topic. It encompasses: 
1) {\it Machine learning expertise} for processing multi-modal sensory inputs, learning control policies, and performing high-level planning;
2) {\it Robotics design principles} for modeling and optimizing the robot’s dynamics and kinematics; and
3) {\it Mechanical engineering knowledge} for the manufacturing of structural components and the integration of hardware systems. Each of these topics represents a significant research field with its own dedicated communities and research groups.
% Traditionally, these research communities have evolved and been explored independently. For instance, machine learning is closely aligned with data-driven AI approaches, often emphasizing theoretical and methodological advancements in software systems. In contrast, robotics design and mechanical engineering emphasize the physical realization of hardware systems. However, when it comes to humanoid robot co-design, it fundamentally relies on the joint optimization of these domains, due to their deep and intricate interdependencies. Its advancement demands interdisciplinary collaboration, and the development of unified frameworks capable of optimizing control algorithms, morphological design, and hardware implementation in a coherent and efficient manner.
% In recent years, cross-disciplinary collaboration, particularly under the AI+"X" paradigm, has played a vital role in advancing scientific and technological breakthroughs. For example, AlphaFold~\cite{abramson2024accurate} exemplifies the synergy between deep neural networks and structural biology, revolutionizing protein structure prediction. OpenAI Five~\cite{berner2019dota} and AlphaStar~\cite{vinyals2019grandmaster} integrate deep RL with the gaming and entertainment industry, pushing the boundaries of AI in complex, multi-agent environments. 
% Similarly, Med-PaLM~\cite{singhal2025toward} bridges LLMs with medical knowledge, enabling AI-assisted healthcare solutions. These successes highlight the transformative potential of combining AI with domain-specific expertise. 
In this context, we argue that {\it {humanoid robot co-design provides a shared research agenda for collaboration across AI, robotics, and engineering; this collaboration is an enabling condition rather than evidence of technical necessity.}}

\vspace{-0.15in}
\section{{Toward Physical RSI: Open Research Questions}}\label{sec:open-questions}
\vspace{-0.1in}
To encourage further exploration of humanoid robot co-design, we propose a set of open questions that may be addressed in both the short and long term.

\noindent{\bf Open Questions in Short-Term.} We present open research questions that could be effectively tackled by leveraging emerging techniques and models.

% \guiliang{Efficent Design Representation Learning?}

1) {\it Align Software Design to Hardware Setting.}
Standard robot description formats, such as the Unified Robot Description Format (URDF) and MuJoCo’s XML model (MJCF), primarily represent a robot’s geometric and kinematic structure. In particular, URDF is largely limited to describing links, joints, and inertial properties, and generally does not encode hardware-level details such as actuator power and control interfaces, sensor connectivity and configuration, wiring, communication protocols, or power distribution. How to align the design based on URDF and MJCF to guide the hardware design remains an open question

2) {\it Efficient Representation for Robot Design.}
Deriving concise and informative representations of data has been a key factor in the success of modern machine learning. In robotic learning tasks, recent studies have explored efficient representations for various types of multi-modal data, including language~\cite{mikolov2013Word2Vec}, 2D images~\cite{radford2021Clip}, 3D shapes such as point clouds~\cite{qi2017pointnet} and scenes~\cite{mildenhall2021nerf,kerbl20233Dgaussians}, and tactile information~\cite{zhao2024transferable}.
However, in most of these studies, the robot structure is fixed, and there has been limited exploration of efficient representations for robot morphology. This lack of focus significantly limits the efficiency of learning and adaptation in tasks involving adaptive robot design, particularly those that rely on updating deep neural networks.

3) {\it Benchmarking Robot Co-Design.} In robot co-design, Sim2Real training plays a crucial role by allowing the performance of co-design algorithms to be evaluated in simulated environments before deployment on physical hardware (Section~\ref{subsec:sim2real}). However, unlike other robotic manipulations and location tasks with rich benchmarks~\cite{James2020RLBench,li2022behavior,luo2023fmb,sferrazza2024humanoidbench,xu2024human}, humanoid robot co-design, as a relatively new research area, lacks commonly applied benchmarks.
Instead, prior studies often customize their tasks and environments according to specific goals, making it difficult to assess how well these algorithms generalize to other settings.
While these case-specific studies provide valuable insights for physical robot structure, their applicability to other embodied humanoid robots, especially in diverse tasks and environments, remains uncertain. 
As a result, establishing a standardized benchmark for humanoid robot co-design emerges as an important and timely objective for the field, especially with the availability of simulation platforms such as Isaac Gym~\cite{Makoviychuk2021Isaac}, MuJoCo~\cite{TodorovMuJoCo2012}, and Genesis~\cite{Genesis}.

4) {\it Design-Aware Policy Optimization.} 
In addition to learning meta-policies that can adapt to different tasks and environments, an important open question is how to develop design-aware policies $\pi_\psi:\mathcal{S}\times\mathcal{G}\times \Psi\rightarrow\mathcal{A}$~\cite{fadini2024making}.
Such policies are designed to generalize effectively across a range of different robot morphologies, enabling adaptive action control even when the physical structure $\psi$ changes. 
When a modification in the robot’s body occurs, the policy can still perform reasonably well and, with minimal fine-tuning, adapt to the new structure. In this way, the policy can effectively serve as a morphology-aware controller, reducing the need for retraining from scratch whenever structural changes are introduced. This capability is crucial for co-design frameworks, where iterative updates to both control and morphology are expected. Ultimately, robust generalization across morphologies not only accelerates the Sim2Real transfer process but also enhances the practicality and scalability of humanoid robot deployment in dynamic, real-world environments.

\noindent\textbf{Long-term open questions.}

\textit{1) Can the field build shared digital twins and morphology benchmarks for humanoids?}
Without shared simulation assets, cost models, and evaluation protocols, humanoid co-design will remain accessible only to a few well-funded organizations.
A useful benchmark should evaluate not only task success, but also energy consumption, hardware cost, policy retraining cost, robustness, safety, and cross-task generality.
Such shared infrastructure would make it possible to compare fixed-body learning, morphology-only optimization, and full body-control co-design under a common protocol.

\textit{2) Can humanoid co-design be made safe and certifiable?}
Changing morphology changes failure modes. Long-term progress requires methods that provide safety guarantees, validate contact-rich behavior, and track risk introduced by hardware changes.
% For example, longer limbs, heavier payload modules, or more compliant joints may improve capability while also altering balance, collision risk, actuator stress, and fall recovery.
A key challenge is therefore to develop safety-aware co-design methods that can screen candidate morphologies before deployment.

\textit{3) Can we learn foundation policies that span families of humanoid bodies?}
A long-term goal is training a policy model across many morphologies and tasks, allowing new humanoid variants to be evaluated and adapted with limited additional data.
Such a policy should condition not only on observations and goals, but also on morphology descriptors such as link lengths, joint limits, actuator capacities, and end-effector types.
This would reduce the need to retrain a controller from scratch whenever the robot body changes, making iterative co-design substantially more practical.

\textit{4) Can rapid fabrication or reconfigurable hardware reduce the cost of physical validation?}
The bottleneck in humanoid co-design is not only optimization but also hardware iteration. Progress in modular actuators, adjustable limbs, rapid prototyping, and standardized interfaces could make physical co-design more broadly accessible.
Instead of rebuilding an entire humanoid for every design candidate, researchers could test bounded morphology changes through swappable hands, tunable compliance, movable sensors, or adjustable mass distributions.
Such reconfigurable platforms would provide a practical bridge between simulation-only co-design and full hardware redesign.

{\noindent\textit{5) Can co-design systems improve their own improvement mechanisms?}
Beyond optimizing successive bodies and policies, physical RSI requires learning from previous design cycles to improve design representations, candidate selection, and controller adaptation. Progress should be measured by whether accumulated experience reduces the resources needed to discover better body--policy combinations on held-out design problems. Improvements must remain subject to engineering constraints and physical validation.}

\vspace{-0.1in}
\section{Conclusion}
\vspace{-0.1in}
% This paper advocates for a co-design paradigm in humanoid robotics, where both control policies and physical morphology are jointly optimized. Inspired by biological evolution, we argue that co-design is not only feasible, with techniques like strategic exploration, Sim2Real learning, and meta-policy control, but also necessary for achieving embodied intelligence.
% By addressing key challenges and outlining open research directions, we position co-design as a foundational approach for developing intelligent, adaptable, and general-purpose humanoid robots capable of thriving in complex real-world environments.
This paper advocates for a body-control co-design paradigm in humanoid robotics, emphasizing the joint optimization of both control strategies and physical morphology. Inspired by principles of biological evolution, {we argue that co-design deserves a central role when physical embodiment limits the capabilities required for real-world tasks.}
{We discuss potential routes toward feasibility through strategic exploration, Sim2Real transfer, and meta-policy learning, and examine the methodological, application-driven, and interdisciplinary motivations for co-design.} 
By integrating co-design into the development pipeline, we can embrace the potential for more generalizable and intelligent humanoid systems.
{Together, these directions position humanoid co-design as a path toward physical RSI: improving bodies and policies through experience, while progressively improving the mechanisms that guide their development.} 
% We position co-design as a foundational approach for developing intelligent, adaptable, and general-purpose humanoid robots capable of thriving in complex real-world environments.

% In the unusual situation where you want a paper to appear in the
% references without citing it in the main text, use \nocite
% Removed unresolved template-only nocite{langley00}.

% \newpage
\section*{Acknowledgments} This work is supported in part by NSFC under Grant 62606447,
Shenzhen Science and Technology Program under grant KJZD20240903104008012, Shenzhen Science and Technology Program under grant ZDCY20250901113000001, CUHK-CUHK(SZ)-GDSTC Joint Collaboration Fund No. 2025A0505000053, 
and GuangDong Key Laboratory of Big Data Computing (2021B1212040002).

\bibliography{bibliography}
\bibliographystyle{unsrt}

%%%%%%%%%%%%%%%%%%%%%%%%%%%%%%%%%%%%%%%%%%%%%%%%%%%%%%%%%%%%%%%%%%%%%%%%%%%%%%%
%%%%%%%%%%%%%%%%%%%%%%%%%%%%%%%%%%%%%%%%%%%%%%%%%%%%%%%%%%%%%%%%%%%%%%%%%%%%%%%
% APPENDIX
%%%%%%%%%%%%%%%%%%%%%%%%%%%%%%%%%%%%%%%%%%%%%%%%%%%%%%%%%%%%%%%%%%%%%%%%%%%%%%%
%%%%%%%%%%%%%%%%%%%%%%%%%%%%%%%%%%%%%%%%%%%%%%%%%%%%%%%%%%%%%%%%%%%%%%%%%%%%%%%
\newpage
\appendix
\onecolumn
\section{A Summary of Representative Studies in Robotic Co-Design}

Co-design for embodied agents has emerged as an important paradigm for developing intelligent systems. ~\cite{wang2025embodied} summarizes recent advances in the design of agents ranging from virtual creatures to physical robots. Since this study focuses on justifying the rationale for co-designing humanoid robots rather than providing a comprehensive review, Table~\ref{tab:summary-co-design} presents a selection of representative recent progress in robotic co-design.
{The table is a selected overview rather than an exhaustive survey of LLM-assisted robot design; generating candidate structures and evaluating them through coupled controller adaptation are related but distinct contributions.}
{These directions are complementary: foundation models may propose candidate designs, while simulation and controller adaptation evaluate their behavioral consequences. Our focus is the closed-loop coupling between design and control, rather than structure generation alone.}
We observe that most of these works focus on relatively simple robot types, such as modular and quadrupedal robots, with a limited number of design parameters considered.
More importantly, the majority of these studies are tailored to specific tasks and environments. Extending these co-design methods to more complex humanoid robots operating across a diverse range of tasks and settings remains an important but largely unexplored challenge.
\begin{table}[htbp]
    \centering
    % \vspace{0.1in}
    \caption{The summary of recent studies and progress in robotic co-design. We list the studies that are partially related to humanoid robot co-design after the dashed line.}
    \renewcommand{\arraystretch}{1.4}
    \resizebox{\textwidth}{!}{
    \begin{tabular}{l|c|l}
    \specialrule{1pt}{0pt}{-1pt}
    \textbf{Designing Method}  & \textbf{Robot Type} & \textbf{Designing Parameters} \\
    \specialrule{1pt}{0pt}{-1pt}
    Meta RL~\cite{Belmonte2022RL4Leggy} & Quadrupedal Robot & Thigh and shank lengths; gear ratios of the actuators. \\ %\hline
    Bayesian Optimization~\cite{Bjelonic2023CoDesignQuad} & Quadrupedal Robot & Parameters of parallel elastic knee joint. \\ %\hline
    ADMM~\cite{bravo2024engineering} & Quadrupedal Robot & Parameters of parallel elastic actuation (PEA). \\ %\hline
    Bayesian Optimization~\cite{chen2024pretraining} & Quadrupedal Robot & Thigh and shank lengths. \\ %\hline
    Implicit Function Theorem~\cite{ha2017joint} & Quadrupedal Robot & Link length; actuator poses. \\ %\hline
    Adjoint Method~\cite{desai2018interactive} & Quadruped and Hexapod Robot & Link lengths; actuator poses; robot's width and length \\ %\hline
    PPO~\cite{Li2024RL} & Modular Soft Robot & 3D voxel-wise material assignments and spatial placement.  \\ %\hline
    LLM-aided Evolution Search~\cite{song2025laser} & Modular Soft Robot & 3D voxel-wise material assignments and spatial placement.\\ %\hline
    Model Order Reduction~\cite{Schaff2022SoftRLDesign} & Modular Soft Robot & The combination of actuator placement and pressure regulators. \\ %\hline
    DQN~\cite{Whitman2020RLDesign}  & Modular Manipulating Robot & The combination of different modules. \\ %\hline
    RoboGAN~\cite{hu2022modular} & Modular Locomoting Robot & Module type assignment on fixed-topology graph.\\ %\hline
    Graph Neural Network~\cite{iii2021taskagnostic} & Modular Locomoting Robot & Size and position of limbs, type and range of joints \\ %\hline
    % Transformer~\cite{kurin2021my} & Modular Locomoting Robot & Size and position of limbs, and type and range of joints. \\ \hline
    Particle Swarm Optimization~\cite{Luck2019CoAdapt} & Modular Locomoting Robot & Leg segment lengths. \\ %\hline
    PPO~\cite{pathak2019learning} & Modular Locomoting Robot & The combination of limbs. \\ %\hline
    Neural Graph Evolution~\cite{wang2018neural} & Modular Locomoting Robot & The combination of different modules. \\ %\hline
    PPO~\cite{lu2025bodygen} & Modular Locomoting Robot & Limb length and size; joint rotation range and torque limit. \\ %\hline
    Quadratic Programming~\cite{spielberg2017functional} & Ariticulated Robot & Geometry and inertia of links; torque limits of joints. \\ %\hline
    PPO~\cite{Schaff2019RLDesign} & Legged Locomoting Robot & Link length and mass. \\ %\hline
    CMA-ES~\cite{li2025generating} & Freedom Endoskeletal Robot & Limb length; soft and rigid radii. \\ %\hline
    DGDM~\cite{Xu2024DGDM} & Sensor-less Jaw Manipulator & Manipulator finger geometry (represented as Bézier curves). \\ %\hline
    Binary Programming~\cite{Carlone2019CoDesign} & Autonomous Racing Drone & The combination of different modules. \\\hdashline
    Evolution RL~\cite{Cheng2024SERL} & Lightweight Bipedal Robot & Thigh and shin lengths. \\ %\hline
    HZD Optimization~\cite{Ghansah2023HumanoidCoDesign} & Bipedal Robot & Thigh and shin lengths. \\ %\hline
    Genetic Algorithm~\cite{Sartore2023HumanoidCoDesign} & ErgoCub2 Humanoid Robot & Motor types and link inertial parameters. \\ %\hline
    ES/RL/CMA-ES/Bayesian Optimization~\cite{yue2025toward} & Humanoid Robot & Size, mass, joint stiffness, and damping. \\
    CMA-ES~\cite{jin2025evolutionary} & RH5 Humanoid Robot & Gear ratio of lower limbs, shoulder, elbow, and wrist actuators. \\
    Particle Swarm Optimization~\cite{du2025efficient} & Kuafu Humanoid Robot & Thigh and shank lengths. \\
    Gradient-based Optimization~\cite{sartore2022optimization} & iCub Humanoid Robot & Links geometry and density. \\
    Spectral Design Evolution~\cite{sun2026evolving} & Musculoskeletal Robot & Strength, velocity, and stiffness of symmetry-grouped muscles.\\
    \specialrule{1pt}{0pt}{0pt}
    \end{tabular}
    }
    \label{tab:summary-co-design}
\end{table}

\begin{figure}[htbp]
    \centering
    % \vspace{-0.5in}
    \includegraphics[width=0.7\linewidth]{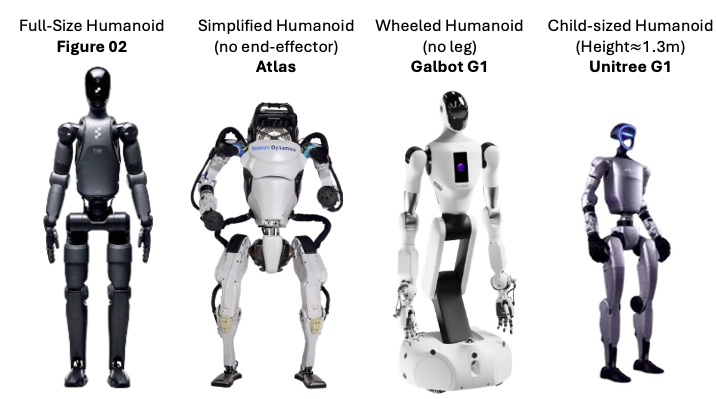}
    % \vspace{-0.2in}
    \caption{Examples of humanoid robots with varying physical structures are shown from left to right: full-sized, simplified, wheeled, and child-sized humanoid robots.}
    \vspace{-0.2in}
    \label{fig:humanoid-examples}
\end{figure}

% \section*{No checklist is needed for position paper track (See Position Paper Track FAQ).}

%%%%%%%%%%%%%%%%%%%%%%%%%%%%%%%%%%%%%%%%%%%%%%%%%%%%%%%%%%%%%%%%%%%%%%%%%%%%%%%
%%%%%%%%%%%%%%%%%%%%%%%%%%%%%%%%%%%%%%%%%%%%%%%%%%%%%%%%%%%%%%%%%%%%%%%%%%%%%%%

\end{document}